\documentclass[runningheads]{llncs}

\usepackage{eccv}
\usepackage{marvosym}

\usepackage{eccvabbrv}

\usepackage{graphicx}
\usepackage{booktabs}
\usepackage{tabularx}
\usepackage{ragged2e} 

\usepackage{multirow}
\usepackage{array}
\usepackage{makecell}
\usepackage{caption}
\usepackage{algorithm}
\usepackage{amsmath}
\usepackage{algpseudocode} 
\usepackage{amssymb}

\usepackage[accsupp]{axessibility}  

\usepackage{graphicx}
\usepackage{booktabs}
\usepackage{tabularx}
\usepackage{ragged2e} 
\usepackage{array}
\usepackage{makecell}
\usepackage{caption}
\usepackage{algorithm}
\usepackage{amsmath}
\usepackage{algpseudocode} 
\usepackage{amssymb}
\usepackage{tcolorbox}       
\usepackage{listings}        
\usepackage{xcolor}
\usepackage{caption}
\usepackage{makecell}

\usepackage{hyperref}
\usepackage{makecell} 

\usepackage{orcidlink}

\begin{document}

\title{NeuroRefiner: Morphology-Aware Multi-Agent Refinement for 3D Fluorescence Microscopy Neuron Segmentation} 

\titlerunning{NeuroRefiner}


\author{Haiyang Yan\inst{1,2}\orcidlink{0009-0003-7296-7034} \and
Jinyue Guo\inst{1,3}\orcidlink{0009-0007-4042-8182} \and
Yanchao Zhang\inst{1,2}\orcidlink{0009-0008-9313-5815} \and
Bingqing Wang\inst{1,3}\orcidlink{0009-0007-9517-0182} \and
Zhenchen Li\inst{4}\orcidlink{0000-0002-5015-3529} \and
Jing Liu\inst{1}\orcidlink{0000-0002-8386-9187} \and
Jiazheng Liu\inst{1}\orcidlink{0000-0002-2001-1597} \and
Linlin Li\inst{1}\orcidlink{0000-0003-3974-4130} \and
Hua Han\inst{1,2}\textsuperscript{(\Letter)}\orcidlink{0000-0003-4713-4631}}

\authorrunning{H.~Yan et al.}

\institute{State Key Laboratory of Brain Cognition and Brain-inspired Intelligence Technology, Institute of Automation, Chinese Academy of Sciences\\ 
\email{\{yanhaiyang2022, hua.han\}@ia.ac.cn}\and
School of Future Technology, University of Chinese Academy of Sciences \and
School of Artificial Intelligence, University of Chinese Academy of Sciences \and
The State Key Laboratory of Cognitive Neuroscience and Learning, Beijing Normal University, Beijing, China
}

\maketitle

\begin{abstract}
Accurate 3D neuron segmentation in fluorescence microscopy is critical for neuroscience. However, the sparse and elongated morphology of neurons poses significant challenges to existing segmentation methods. These methods struggle to preserve both local details and global topology, leading to fragmented results. To address this, we propose NeuroRefiner, a multi-agent system that formalizes the human expert workflow involving iterative global observation and local editing. Specifically, NeuroRefiner comprises three collaborative agents dedicated to diagnosing topological errors, generating correction instructions, and validating refinement quality. To facilitate agent instruction-guided segmentation refinement, we propose TopoRefineNet, a dedicated 3D U-Net-based tool that leverages cross-modality feature fusion to generate refined masks. Through multi-round agent reasoning and voxel-level editing, NeuroRefiner produces topologically more accurate segmentations with enhanced interpretability. Experiments on the BigNeuron, CWMBS, and ZBFWB datasets demonstrate that NeuroRefiner outperforms state-of-the-art methods, notably achieving a 3.02\% improvement in F1 score on the challenging ZBFWB dataset.

  \keywords{Neuron reconstruction \and Multi-agent system \and Segmentation refinement \and Fluorescence Microscopy}
\end{abstract}

\section{Introduction}
\label{sec:intro}

Neuronal reconstruction aims to derive neuronal skeletons from 3D fluorescence microscopy volumes, thereby enabling quantitative analysis in neuroscience \cite{gou2024gapr,zhang2023collaborative}. Accurate reconstruction requires not only recovering fine-grained structural details but also preserving topological correctness. 
Unlike electron microscopy (EM) neuron reconstruction \cite{sheridan2023local,jiang2026neuromamba,zhang2024segneuron}, which typically deals with dense cellular structures, fluorescence microscopy poses distinct challenges. These challenges stem from extreme signal sparsity and filamentous morphology within high-resolution 3D volumes that are frequently corrupted by substantial noise \cite{liu2022neuron,chen2023deep}. Consequently, robust automated reconstruction requires multi-scale features to preserve both structural details and global connectivity.

\begin{figure*}[!t]
\centerline{\includegraphics[width=1\columnwidth]{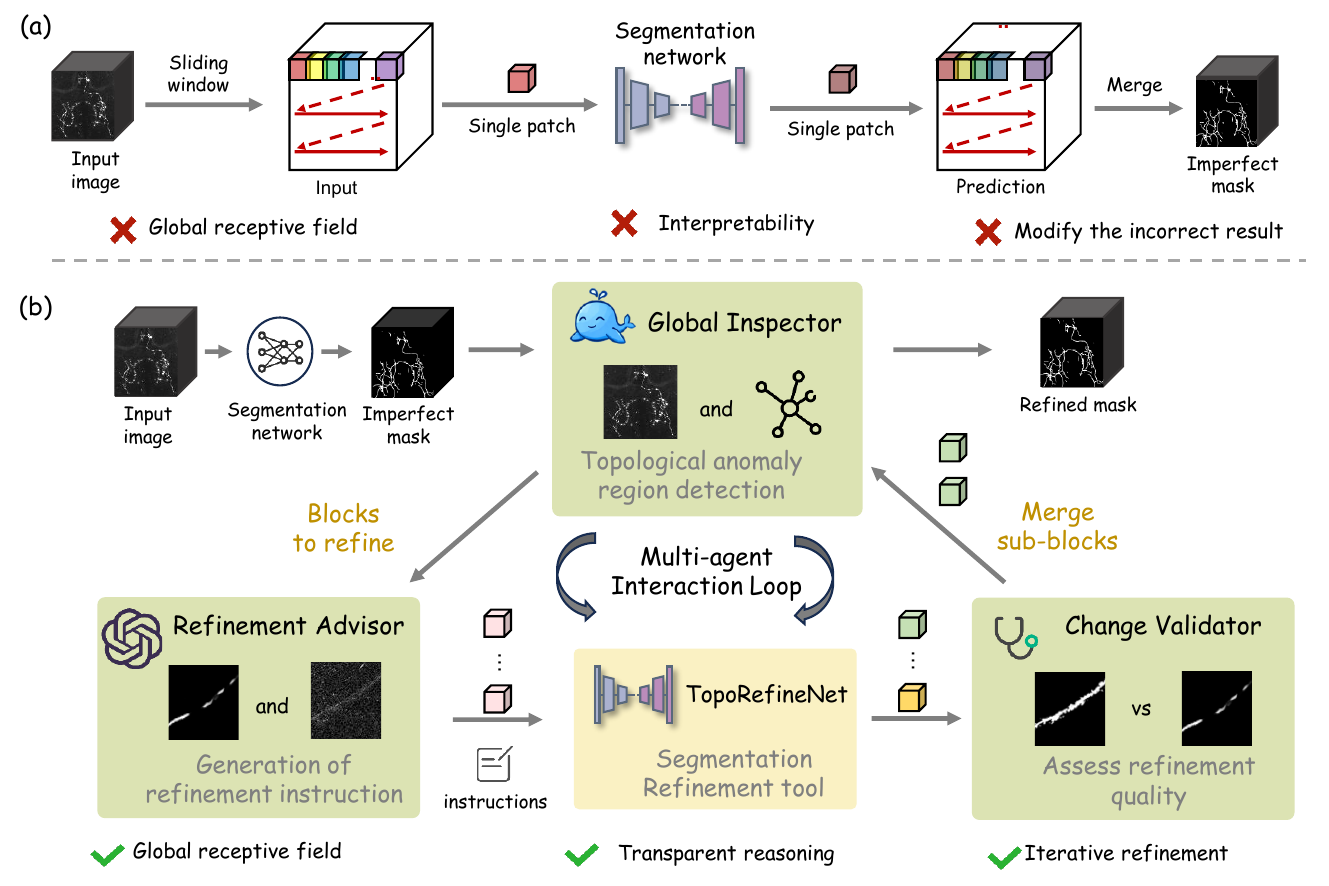}}
\caption{(a) Mainstream neuron segmentation paradigms. Sliding window-based methods fail to preserve holistic neuronal morphology. Moreover, a single forward pass cannot rectify segmentation errors and lacks interpretability.
(b) Overview of NeuroRefiner. By combining global topological observation with local fine-grained refinement, it generates continuous segmentation results consistent with topological priors. The explicit and transparent reasoning process of the agents offers enhanced interpretability.}
\label{fig1}
\end{figure*}

Neuron segmentation plays a pivotal role in neuronal reconstruction by enhancing weak signals and suppressing noise \cite{liu2025netracer,liu2024brain}. As shown in \cref{fig1} (a), given the prohibitive size of raw image volumes, existing segmentation approaches \cite{wang20213d,yan2024neurolink,yang2021structure,wang2023nrtr} typically rely on sliding-window strategies for inference. Nevertheless, the limited receptive field in each local window precludes the modeling of long-range dependencies essential for preserving global neuronal morphology \cite{yan2025glancing}, resulting in fragmented segmentation. Furthermore, these end-to-end approaches typically operate as black boxes, lacking interpretability and offering no mechanism for the correction of significant topological errors in segmentation. Such errors are challenging to localize and rectify, ultimately constituting a critical bottleneck for the automation of neuronal reconstruction.
 
Agent-based biomedical image segmentation methods \cite{jiang2025incentivizing,yu2025gencellagent,jiang2026ibisagent} have demonstrated enhanced performance and interpretability via the reasoning and tool invocation capabilities of large language models (LLMs). However, existing agents and their associated tool designs remain ill-suited for neuron segmentation. First, they depend on foundation models (e.g., MedSAM2 \cite{ma2025medsam2}) as tools, which process 3D volumes slice-by-slice using 2D encoders. Yet, this proves inadequate for neurons, as their structural sparsity and ambiguous boundaries hinder effective feature extraction within isolated 2D planes \cite{zhang2023towards}.
Furthermore, these agents confine the workflow to single tool invocations, which lack the flexibility to iteratively refine the segmentation via agent instructions. These limitations underscore the critical need for a dedicated agent framework equipped with domain-specific tools tailored for neuron segmentation.

Motivated by the neuroscientist's cognitive paradigm of global observation and local refinement, we propose NeuroRefiner, a novel multi-agent framework designed for iterative, topology-aware neuron segmentation. As depicted in \cref{fig1} (b), our approach comprises three specialized agents: the Global Inspector, which leverages initial segmentation masks and their corresponding Betti numbers to pinpoint sub-regions requiring correction; the Refinement Advisor, which generates detailed refinement instructions for each sub-region; and the Change Validator, which assesses refinement validity and determines whether to accept or regenerate the refinement instructions. Complementing these reasoning agents is a dedicated execution tool, TopoRefineNet, which performs instruction-guided editing of the segmentation mask. This closed-loop pipeline effectively leverages multi-scale features and operates iteratively until the segmentation satisfies the structural completeness of neurons. In summary, our contributions are threefold:\begin{itemize}
\item To overcome the limitations of existing neuron segmentation approaches, we present NeuroRefiner—the first LLM-based multi-agent system that achieves topology-aware iterative optimization of segmentation results.

\item  To enable efficient collaboration between reasoning agents and specialized segmentation models, we design TopoRefineNet, a dedicated segmentation editing tool, together with a tailored two-stage training strategy that aligns model behavior with structural refinement instructions.

\item Experiments on three benchmarks demonstrate that our method outperforms existing neuron segmentation and segmentation refinement approaches.
 
\end{itemize}

\section{Related Work}

\subsubsection{Neuron Segmentation.} Recent advancements have prioritized the design of specialized architectures to extract multi-scale features, aiming to simultaneously model global neuronal morphology and fine-grained structural details. Architecturally, this is typically realized through the integration of dedicated convolution \cite{yan2024neurolink,yang2020neuron} or graph-based reasoning modules \cite{wang2021single}. In terms of morphological constraints, methods such as SGSNet \cite{yang2021structure} and Tubular \cite{wang20213d} augment feature representation via multi-task learning that predicts auxiliary geometric cues. To capture intrinsic morphological priors, MP-NRGAN \cite{chen2021weakly} employs adversarial training utilizing weak supervision signals derived from reconstructed neurons. Furthermore, to mitigate the limitations of restricted receptive fields inherent in patch-based inference, GBP-Net \cite{yan2025glancing} encodes long-range contextual information and fuses it with intra-patch features.

Despite these advancements, prevailing models remain hindered by single-pass inference paradigms, which preclude iterative error correction. While certain approaches \cite{2019Learning,zhao2023pointneuron} utilize point cloud networks to refine segmentation outputs, they fundamentally fail to recall omitted structural segments (false negatives) missed during the initial prediction. Conversely, our framework introduces a multi-round refinement mechanism guided by multi-scale features, effectively mitigating the errors in single-pass models.

\vspace{-1em}

\subsubsection{Agent for Biomedical Image Segmentation.}
Existing agent-based approaches generally fall into two distinct paradigms. The first paradigm relies on conventional multi-agent reinforcement learning (MARL) for low-level, pixel-wise decision-making. Methods such as IteR-MRL \cite{liao2020iteratively} and BS-IRIS \cite{ma2020boundary} employ multi-agent reinforcement learning for interactive 3D segmentation, treating voxels as collaborative agents to iteratively refine results. Similarly, dbMiM \cite{chen2023self} employs MARL to optimize masking strategies for self-supervised neuron segmentation. However, these approaches are predominantly tailored for electron microscopy or general 3D modalities (e.g., MRI/CT), rendering them ineffective for handling neurons in fluorescence microscopy.

The second paradigm represents a more advanced, LLM-driven approach. These agents are LLM-centric and leverage planning-execution loops to invoke external segmentation tools. For instance, Ophiuchus \cite{jiang2025incentivizing} adopts a three-stage training strategy to coordinate vision tools like SAM2 \cite{2024SAM} and BiomedParse \cite{2024BiomedParse} for enhanced lesion segmentation. IBISAgent \cite{jiang2026ibisagent} generates textual click instructions to invoke interactive segmentation tools for progressive mask refinement, while GenCellAgent \cite{yu2025gencellagent} introduces a multi-agent framework that achieves intelligent tool routing via a planner-executor-evaluator loop. Although these methods have improved accuracy, they lack topology-aware agent design and tools specifically tailored for sparse neurons.

\vspace{-1em}
\subsubsection{Segmentation Refinement.} High-resolution image segmentation conventionally necessitates downsampling operations or patch-based inference strategies, which incur boundary artifacts or information loss. To mitigate these issues, a series of refinement methods has been proposed. Mask Transfiner \cite{ke2022mask} identifies incoherent regions to perform fine-grained label correction. SegRefiner \cite{wang2023segrefiner} introduced a generic, iterative framework leveraging discrete diffusion processes to correct diverse segmentation errors. To achieve accurate segmentation of EM neurons, FGNet \cite{li2025fgnet} refines the semantic features extracted by SAM2 \cite{ravi2024sam} by training a lightweight fine-grained encoder.
While effective for boundary regularization, existing methods neglect global topology, which is paramount for preserving filamentous neuronal structures. Hence, topology-aware refinement for neuronal structures remains a critical, underexplored frontier.

\begin{figure*}[!t]
\centerline{\includegraphics[width=1\columnwidth]{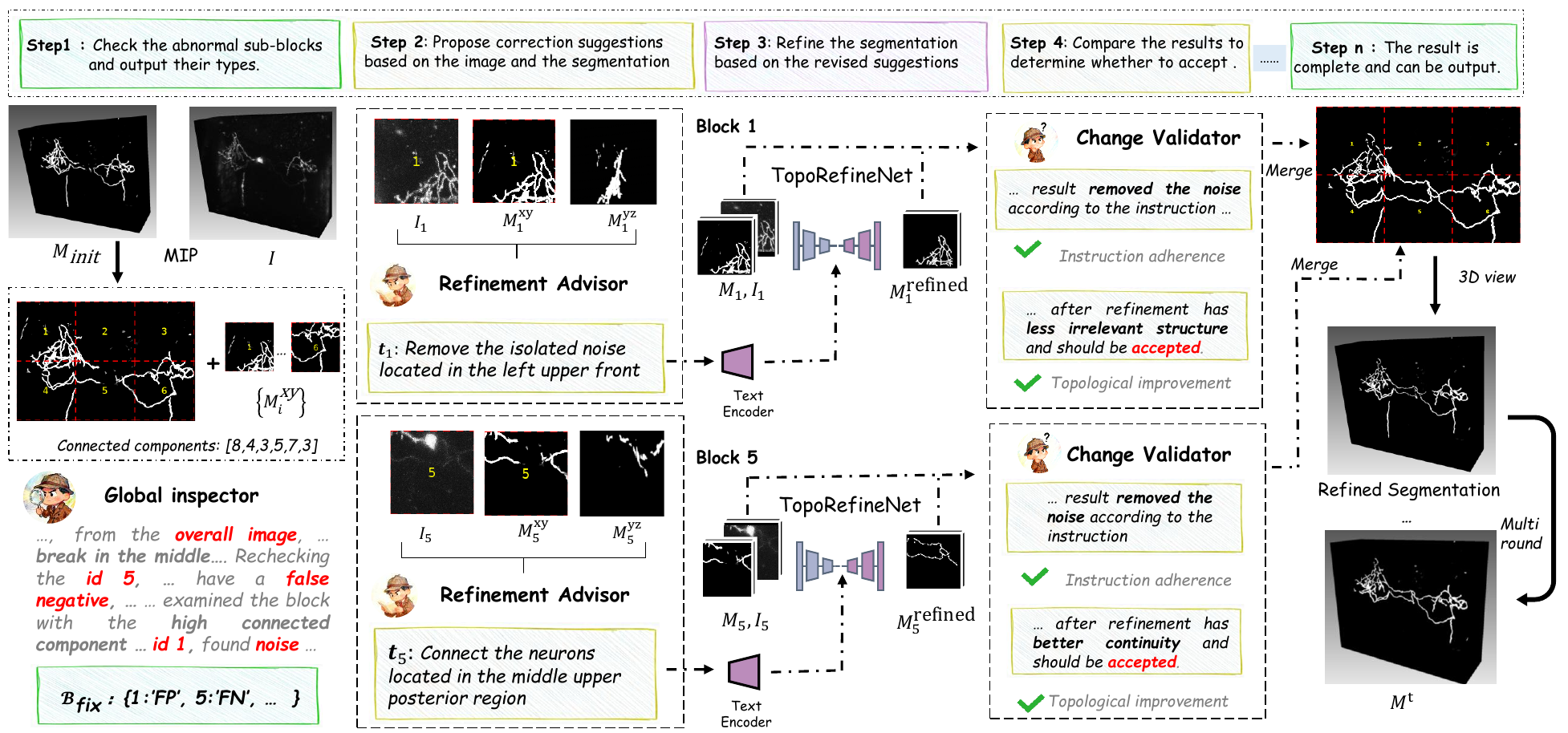}}
\caption{Collaborative NeuroRefiner workflow. The system performs multi-round refinement via the interaction of agents and TopoRefineNet to obtain topologically complete segmentation results.}
\label{fig2}
\end{figure*}

\section{Method}
\cref{sec:refine} describes the collaborative mechanism among agents in NeuroRefiner. \cref{sec:agents} details the design of each individual agent and \cref{sec:unet} presents the architecture and training strategy of our proposed tool, TopoRefineNet.

\begin{algorithm}
\caption{NeuroRefiner: Morphology-Aware Multi-Agent Refinement}
\label{al1}
\begin{algorithmic}[1]
\Require 3D image $I$, initial mask $M^{(0)}$, max iterations $T_{max}$, iteration $t \gets 0$
\While{$t < T_{max}$ }
    \State Compute z-axis MIP $M^{xy}$, sub-blocks $\{b_i\}$, and Betti numbers $\{\beta_0(b_i)\}$
    \State $B_{fix}, \{\tau_i\} \gets \pi_{ins}(M^{xy}, \{M_i^{xy}\}, \{\beta_0(b_i)\})$ // Identify blocks to fix
    \State if $B_{fix}=\varnothing$ then \textbf{break} //morphologically converged
    \For{each block $b_i \in B_{fix}$}
        \State Initialize retry counter $k \gets 0$
        \While{$k < T_{max}$}
            \State $t_i \gets \pi_{adv}(I_i^{xy}, I_i^{yz}, M_i^{xy}, M_i^{yz}, \tau_i)$ // Generate refinement instruction
            \State $M_i^{refined} \gets \text{TopoRefineNet}(I_i, M_i, t_i)$ // Instruction-guided editing
            \State $r_i \gets \pi_{val}(M_i^{xy}, M_i^{\text{refined}}, t_i, \tau_i)$ // Validate refinement quality
            \If{$r_i$ is Accept}
                \State $M[b_i] \gets M_i^{\text{refined}}$, break //Update local mask
            \Else
                \State  $k \gets k + 1$ //Retry
            \EndIf
        \EndWhile
    \EndFor
    \State $t \gets t + 1$
\EndWhile
\State \Return $M^{\text{final}} \gets M^{(t)}$
\end{algorithmic}
\end{algorithm}

\subsection{NeuroRefiner}

\label{sec:refine}
Manual refinement of neuron segmentation follows a hierarchical workflow in which experts iteratively alternate between global morphological assessment and fine-grained local editing. Motivated by this paradigm, we propose a multi-agent collaborative framework, as outlined in Algorithm \ref{al1}.
The segmentation mask \(M^{(t)}\) is initialized using an off-the-shelf segmentation model. At each iteration, the Global Inspector conducts a holistic assessment of \(M^{(t)}\) to detect sub-regions that violate topological priors. Subsequently, the Refinement Advisor generates a set of structure-aware refinement instructions $\left\{\mathbf{t}_i\right\}$, each tailored to a specific anomalous sub-region. These instructions guide TopoRefineNet to produce refined sub-region masks, denoted as $\{M_i^{\mathrm{refined}}\}$. As a validation mechanism, the Change Validator acccepts only those refinements that yield an improvement over the original segmentation, as

\begin{equation}
M^{(t+1)}  = 
\text{Merge}\left( M^{(t)}, \left\{ M_i^{\text{refined}} \mid \text{Validator}(M_i^{\text{refined}}) = \text{True} \right\} \right).
\label{eq:important}
\end{equation}

As illustrated in \cref{fig2}, the process terminates when the Global Inspector determines that the neuronal morphology is complete or upon reaching \(T_{\text{max}}\) iterations. This iterative protocol ensures transparency, as the explicit diagnostic decisions and instructions generated at each step are auditable.

\subsection{Design of Agents in NeuroRefiner}
\label{sec:agents}

\subsubsection{Global Inspector} leverages the high discriminability of global morphological anomalies to identify sub-regions violating topological priors. Given an initial 3D segmentation mask $M_{init}$ and the corresponding image $I$, we first apply Maximum Intensity Projection (MIP) along the z-axis to obtain projection $M^{xy}$ and $I^{xy}$. Performing MIP on sparse volumes facilitates their input into VLMs while inducing minimal signal overlap.
The projected volume is then partitioned into a set of non-overlapping 2D sub-blocks $\mathcal{B} = \{b_i\}_{i=1}^N$, yielding corresponding image and mask patches $\left\{I_i^{xy}\right\}$ and $\left\{M_i^{xy}\right\}$. 

As a crucial metric for neuron segmentation, topological correctness serves as an effective guide for identifying erroneous regions. For each sub-block, the 0th-order Betti number is computed to quantify topological irregularities as $\beta_0(b_i)=CC(M_i)$, where $CC(\cdot)$ denotes the connected component counting function in the corresponding 3D region. An elevated $\beta_0$ value, which indicates a greater number of isolated fragments within a sub-block, suggests a higher likelihood of topological errors. Based on this topological metric and the mask, the inspection policy $\pi_{\mathrm{ins}}$ identifies abnormal regions and categorizes their error types:

\begin{equation}
\operatorname{B}_{\mathrm{fix}},{\tau_i}=\pi_{\mathrm{ins}}(M^{xy},\{M_i^{\mathrm{xy}}\},{\beta_0(b_i)}),
\end{equation}

where $\operatorname{B}_{fix}\subseteq\mathcal{B}$ denotes the subset of blocks requiring refinement and $\tau_i$ labels each block as either false-positive or false-negative. They provide guidance for the Refinement Advisor to generate detailed instructions.

\subsubsection{Refinement Advisor} generates structured, morphology-aware instructions to guide the correction of identified sub-blocks. To precisely delineate the abnormal region along the z-axis, the agent first applies MIP to the 3D region corresponding to $ b_i$ along the x-axis and then localizes the defect in the yz-plane. Upon obtaining the xy and yz plane projections of the target region, it jointly analyzes the image and mask from orthogonal views to uncover connectivity cues and formulate a refinement instruction, as
\begin{equation}
\mathbf{t}_i = \pi_{\mathrm{adv}}\big(I_i^{xy}, I_i^{yz}, M_i^{xy}, M_i^{yz}, \tau_i\big).
\end{equation}
Each instruction $\mathbf{t}_i$ explicitly encodes both a spatial location (e.g., “upper-left-posterior”) and an editing operation (e.g., “connect” or “remove”). The editing module TopoRefineNet subsequently performs refinement based on the instruction and outputs $M_i^{\mathrm{refined}}$. This design decouples reasoning from voxel-level operations by using natural language as an intermediary, thereby fully unleashing the agent’s reasoning and expressive capabilities.

\subsubsection{Change Validator} ensures the reliability of iterative refinement by evaluating the output of TopoRefineNet and preventing erroneous edits from being merged. Specifically, it compares the initial mask $M_i$ against the refined result $M_i^{\mathrm{refined}}$, conditioned on the instruction $\mathbf{t}_i$ and the error type $\tau_i$
to assess whether the edit is beneficial to morphology. This evaluation is formalized as

\begin{equation}
r_i = \pi_{\mathrm{val}}\big(M_i^{xy}, M_i^{\mathrm{refined}}, \mathbf{t}_i, \tau_i\big).
\end{equation}
The Change Validator only approves an edit when two criteria are satisfied: (1) instruction adherence—the modification correctly implements the spatial and operational intent of $\mathbf{t}_i$, and (2) topological improvement—the refined mask exhibits reduced fragmentation or noise without introducing new artifacts. If either condition fails, it triggers the refinement advisor to generate a new instruction, thereby closing the loop with corrective feedback. For cases that remain unresolved after multiple refinement cycles, the system escalates the sub-block to human experts for manual intervention.

\begin{figure*}[!t]
\centerline{\includegraphics[width=1\columnwidth]{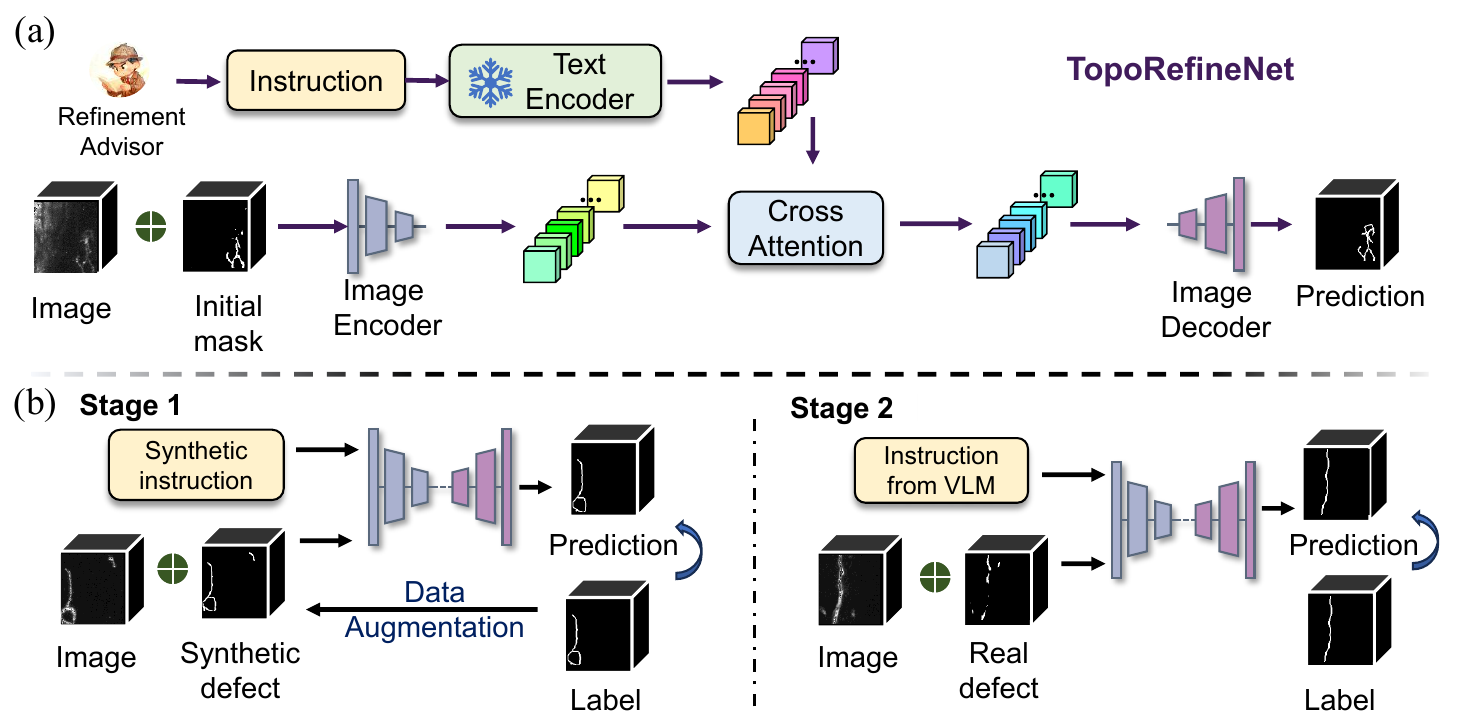}}
\caption{(a) Architecture of TopoRefineNet. The image and instruction are encoded by their respective encoders, fused via cross-attention, and decoded to generate the corrected mask. (b) The proposed two-stage training paradigm. The first stage is trained on synthetic defects, while the second stage utilizes real masks with defects.}
\label{fig3}
\end{figure*}

\subsection{TopoRefineNet}
\label{sec:unet}
\subsubsection{Architecture.}
Serving as the critical bridge between high-level agent reasoning and low-level pixel operations, TopoRefineNet executes the conversion of topological instructions into voxel-level mask modifications. Inspired by recent advances in multimodal conditional generation \cite{zhang2023adding,rombach2022high,couairon2022diffedit}, TopoRefineNet formulates segmentation refinement as an instruction-guided image editing task. As illustrated in \cref{fig3} (a), it takes the image $I_{i}$ and the noisy mask $M_{i}\in\operatorname{R}^{H\times W\times D}$ as input, which are concatenated and encoded by a visual encoder $\mathcal{E}_\theta$ to produce a hierarchical feature pyramid, as
$\{ \mathbf{F}_i^v \}_{i=1}^L = \mathcal{E}_\theta([I_i; M_i])$.
The instruction $\mathbf{t_i}$ is embedded via a frozen pre-trained text encoder, as
$\mathbf{F}^t=\mathcal{T}_\phi(\mathbf{t})\in\operatorname{R}^d.$ 
The interaction is achieved via cross-attention between the deepest visual features $\mathbf{F}_L^v$ and $\mathbf{F}^t$, formulated as
\vspace{-0.5em}
\begin{equation}
\mathbf{F}_L^{\prime v}= \text{CrossAttn}(\mathbf{F}_L^v, \mathbf{F}^t) = \text{softmax}\left(\frac{\mathbf{W}_q \mathbf{F}_L^v (\mathbf{W}_k \mathbf{F}^t)^\top}{\sqrt{d_k}}\right) \mathbf{W}_v \mathbf{F}^t,
\end{equation}
where $\mathbf{W}_q, \mathbf{W}_k, \mathbf{W}_v$ are learnable projection matrices and $d_k$ denotes the feature dimension. This mechanism dynamically recalibrates visual features by attending to regions and editing operations specified in the instruction. The fused feature $\mathbf{F}_L^{\prime v}$ is passed to the decoder $\mathcal{D}$, which integrates it with shallow features via skip connections and upsamples to yield the refined mask $M_{\mathrm{refined}}$. The entire process is concisely expressed as a conditional generation function, as
\begin{equation}
M_{\mathrm{refined}} = \mathcal{F}_\theta(I_i, M_{\mathrm{i}}, \mathbf{t_i}) = \mathcal{D} \circ \text{CrossAttn}\big(\mathcal{E}_\theta([I_i; M_{\mathrm{i}}]), \mathcal{T}_\phi(\mathbf{t_i})\big).
\end{equation}
This design addresses the limitations of general tools in editing sparse neuronal structures, empowering the system with essential capabilities for fine-grained morphological refinement.

\subsubsection{Training Strategy.}
To endow TopoRefineNet with precise instruction-to-voxel editing capability, we devise a two-stage training strategy as shown in \cref{fig3} (b). The initial stage prioritizes cross-modal alignment using synthetic data with morphological defects. Specifically, ground-truth masks \(M_{\mathrm{gt}}\) are perturbed via data augmentation to yield corrupted variants (\cref{details}), with corresponding refinement instructions generated based on the location and type of perturbation. The second stage subsequently shifts focus toward real-world generalization. We curate segmentation outputs from multiple networks trained on diverse neuronal datasets and specifically select instances that exhibit representative topological errors. For each identified error, a refinement instruction is synthesized using Qwen3-VL \cite{bai2025qwen3}. By progressively increasing defect complexity from synthetic scenarios to complex real-world cases, this scheme stabilizes optimization and enhances robustness against the diverse instructions encountered during inference.

\section{Experiments}
\label{sec:blind}

\subsection{Experiments Setup}
\subsubsection{Dataset.}
To evaluate NeuroRefiner across diverse neuronal morphologies and imaging conditions, we conducted experiments on three representative benchmarks, adhering to the training–testing splits established in prior work \cite{yan2025glancing,yang2021structure,liu2024brain}. The BigNeuron \cite{peng2015bigneuron} dataset comprises expert-annotated neurons spanning multiple species. Images were acquired across different laboratories and imaging platforms, exhibiting substantial variations in appearance, resolution, and scale. The CWMBS \cite{liu2024brain} dataset is derived from whole-brain mouse imaging and contains volumes of size 256 × 256 × 256 voxels (physical resolution: 0.2µm × 0.2µm × 1µm), of which 83 exhibit strong background noise and 162 feature thin filamentary structures with weak signals. The ZBFWB dataset \cite{du2025central} consists of confocal microscopy images from 6-day-old zebrafish larvae (1000 × 2000 × 250 voxels; 0.5µm × 0.5µm × 1µm), presenting challenges due to its low contrast, long-range axonal projections, intricate network topologies, and low contrast.

\subsubsection{Metrics.}
To ensure consistency with prior work \cite{yan2025glancing,liu2024brain,yang2021structure}, we use reconstructions generated by APP2 \cite{xiao2013app2} to assess segmentation quality from point-level accuracy to topological completeness. We compute the precision, recall, and F1-score to evaluate voxel-level accuracy and further employ the distance-based metric Spatial Distance (SD) to measure the average bidirectional distance between prediction and ground-truth structures. Its variant SSD \cite{peng2010v3d}, excludes matched points within 2 voxels to focus on significant topological errors. Finally, MES \cite{xie2011anisotropic} evaluates the overall morphological fidelity of topology by quantifying the lengths of missing and extra structures. Lower SD and SSD values, along with higher values for all other metrics, indicate superior reconstruction.

\subsubsection{Details.}
\label{details}
In our experiments, all agents use Qwen3-VL-8B \cite{bai2025qwen3} as the foundation model, and text embeddings are obtained from Qwen3 Embedding 0.6B \cite{zhang2025qwen3}. The Global Inspector performs error detection using a region of 128 × 128 pixels as its fundamental unit. The maximum number of iterations \(T_{\text{max}}\) is set to 5. The prompts for all agents and the template of instructions for TopoRefineNet are provided in the supplementary material.

Both training stages of TopoRefineNet employ cross-entropy loss and dice loss \cite{milletari2016v}. In the first stage, to synthesize FN segmentations, we randomly select a positive sample as the center and perform an erosion operation with a random kernel size $k\in[5,20]$ to generate breaks of varying lengths. For FP samples, we either randomly copy-paste annotations of other neurons or introduce Gaussian noise within local regions. In the second stage, we trained several common architectures—namely, 3D U-Net \cite{cciccek20163d}, V-Net \cite{liu2018improved}, UNETR \cite{hatamizadeh2022unetr}, nnFormer \cite{zhou2023nnformer}, and SwinUNETR \cite{hatamizadeh2021swin}—on a subset of the training set, followed by inference on the complete training set. Based on connected component and F1 scores, we selected a total of 5,310 volumes of 128 × 128 × 64 voxels exhibiting significant segmentation errors for training.

\begin{table}[!t]
\centering
\caption{Quantitative results (\%) on the CWMBS dataset. The best results are \textbf{bolded} and the second-best results are \underline{underlined} in the following table.}
\label{cwmbs}
\small
\setlength{\tabcolsep}{0pt}
\begin{tabular*}{\textwidth}{@{\extracolsep{\fill}}l *{6}{c} *{6}{c}}
\toprule
\multirow{3}{*}{\textbf{Models}} &
\multicolumn{6}{c}{\textbf{Weak Signal}} &
\multicolumn{6}{c}{\textbf{Strong Noise}} \\
\cmidrule(lr){2-7} \cmidrule(lr){8-13}
& SD$\downarrow$ & SSD$\downarrow$ & PRE$\uparrow$ & REC$\uparrow$ & F1$\uparrow$ & MES$\uparrow$ &
  SD$\downarrow$ & SSD$\downarrow$ & PRE$\uparrow$ & REC$\uparrow$ & F1$\uparrow$ & MES$\uparrow$ \\
\midrule

\multicolumn{13}{l}{\textit{\textbf{3D Segmentation Baselines}}} \\
\midrule
{\scriptsize 3D U-Net}  & 28.35 & 33.71 & 93.85&62.37&74.94&46.21 & 24.98&31.61&94.16&69.57&79.23&54.30 \\
{\scriptsize SwinU} &24.94&31.14&94.07&65.07&76.93&56.45 &19.43&27.17&93.71&74.40&80.98&56.30\\
{\scriptsize nnUNet} & 25.97&31.66&95.03&68.82&79.83&54.76 &33.65&39.89&90.38&61.43&73.14&52.19 \\
\midrule
\multicolumn{13}{l}{\textit{\textbf{Neuron-tailored architectures}}} \\
\midrule
{\scriptsize V-Net}  & 23.80&29.97&93.58&71.12&80.82&55.47 &34.96&40.17&88.16&56.34&68.75&46.95 \\
{\scriptsize SGSNet}  & 35.70&40.82&92.44&55.63&69.46&50.29 &19.59&26.79&94.79&73.13&82.56&58.25 \\
{\scriptsize GBP-Net} & 11.85&19.27&95.06&83.50&88.91&67.09 &17.15&24.94&93.41&75.93&83.77&59.41 \\
{\scriptsize ADTL-Net}  & 25.96&31.65&93.49&62.72&75.07&53.51 &28.11&35.18&\underline{95.24}&60.68&74.13&52.79 \\
\midrule

\multicolumn{13}{l}{\textit{\textbf{Refinement approaches on 3D UNet}}} \\
\midrule
{\scriptsize Transfiner}  & 27.33 & 32.14 & 94.10 & 62.28 & 74.95 & 46.22 & 24.53 & 31.52 & 94.33 & 70.22 & 80.51 & 56.28 \\
{\scriptsize SegFix} & 26.74 & 32.09 & 93.00 & 64.72 & 76.32 & 48.93 &23.22&30.18 &\textbf{95.39}&70.11&80.82&57.52 \\
{\scriptsize SegRefiner} & 17.61& 23.79 & 94.92 & 77.84 & 85.54 & 62.31 & 21.64 & 27.80 & 92.19 & 72.84 & 81.38 & 57.91 \\
{\scriptsize Ours} & \underline{9.17} & \underline{15.21} & 95.72 & \underline{85.05} & \underline{90.07} & \underline{70.32} & \textbf{13.29} & \textbf{19.52} & 94.31 & \textbf{79.66} & \textbf{86.37} & \textbf{64.39} \\
\midrule

\multicolumn{13}{l}{\textit{\textbf{Refinement approaches on nnUNet}}} \\
\midrule
{\scriptsize Transfiner}  & 25.34 & 31.73 & 95.63 & 66.29 & 78.30 & 53.16 & 33.24 & 39.70 & 91.69 & 63.79 & 75.24 & 53.27 \\
{\scriptsize SegFix}  & 24.82 & 30.57 & 95.47 & 69.69 & 80.57 & 56.79 & 30.09 & 35.35 & 92.74 & 61.02 & 73.61 & 55.38 \\
{\scriptsize SegRefiner }& 15.98 & 22.60 & \underline{96.71} & 73.46 & 83.50 & 61.41 & 22.53 & 27.83 & 94.35 & 66.37 & 77.92 & 54.93 \\
{\scriptsize Ours} & \textbf{7.26} & \textbf{11.86} & \textbf{96.79} & \textbf{87.39} & \textbf{91.85} & \textbf{72.60} & \underline{15.96} & \underline{20.17}& 93.76 & \underline{77.47} & \underline{84.84} & \underline{61.82} \\
\bottomrule
\end{tabular*}
\end{table}

\subsection{Quantitative Results}
We comprehensively evaluated NeuroRefiner against a diverse set of state-of-the-art (SOTA) methods for fluorescence microscopy neuron segmentation. The baselines include 3D U-Net \cite{cciccek20163d}, SwinUNETR \cite{hatamizadeh2021swin}, and nnUNet \cite{isensee2021nnu}; neuron-tailored architectures SGSNet \cite{yang2021structure}, GBP-Net \cite{yan2025glancing}, Improved V-Net \cite{liu2018improved}, and ADTL-Net \cite{liu2024brain}; refinement-based methods Mask Transfiner \cite{ke2022mask}, SegFix \cite{yuan2020segfix}, and SegRefiner \cite{wang2023segrefiner}. We reproduced the corresponding 3D versions of the 2D-based refinement-based methods using the same training data as TopoRefineNet.

\cref{cwmbs} reports quantitative results on the CWMBS subsets. CNN-based models (3D U-Net, SGSNet) are hampered by local receptive fields, limiting generalization under distribution shifts, whereas multiscale methods (GBP-Net, SwinUNETR) demonstrate superior accuracy. Due to a lack of topology-centric refinement, Transfiner and SegFix yield only marginal gains on the masks of 3D U-Net and nnUNet. In contrast, our approach leverages agent-based reasoning and topology-centric refinement and improves F1 scores of various segmentation masks by over 10\% on the weak-signal subset, establishing SOTA results across both subsets.

\begin{table}[!t]
\caption{Quantitative results (\%) on the BigNeuron and ZBFWB datasets.}
\label{tabzbf}
\small
\centering
\resizebox{0.7\textwidth}{!}{
\begin{tabular}{c | c | c c c c c c}
\toprule
 & Method & \textbf{SD↓} & \textbf{SSD↓} & \textbf{PRE↑} & \textbf{REC↑} & \textbf{F1↑} & \textbf{MES↑} \\
\midrule
\multirow{9}{*}{\rotatebox{90}{BigNeuron}} 
  & 3D U-Net \cite{cciccek20163d} & 7.23 & 11.27 & 83.15 & 89.76 & 86.33 & 66.82 \\
  & nnUNet \cite{isensee2021nnu} & 5.16 & 8.68 & 87.34 & 86.87 & 87.10 & 68.72 \\
  & Improved V-Net \cite{liu2018improved} & 7.40 & 11.16 & 82.50 & \textbf{90.41} & 86.27 & 67.16 \\
  & SGSNet \cite{yang2021structure} & 5.90 & 9.34 & 86.22 & 90.28 & 88.20 & 69.90 \\
  & GBP-Net \cite{yan2025glancing} & 4.89 & 8.41 & 88.83 & 89.23 & 89.03 & 70.77 \\
  & ADTL-Net \cite{liu2024brain} & 5.12 & 8.39 & 88.25 & 87.14 & 87.69 & 69.32 \\
  & Mask Transfiner \cite{ke2022mask} & 6.82 & 10.57 & 86.14 & 88.89 & 87.49 & 69.01 \\
  & SegFix \cite{yuan2020segfix} & 7.18 & 11.35 & 83.92 & 89.19 & 86.47 & 66.32 \\
  & SegRefiner \cite{wang2023segrefiner} & 5.79 & 9.94 & 87.18 & 89.23 & 88.19 & 69.27 \\
  & Ours & \textbf{3.39} & \textbf{6.94} & \textbf{89.74} & 90.37 & \textbf{90.05} & \textbf{72.02} \\
  
\midrule
\multirow{8}{*}{\rotatebox{90}{ZBFWB}} 
  & 3D U-Net \cite{cciccek20163d} & 38.65 & 45.41 & 86.33 & 63.15 & 72.94 & 53.63 \\
  & nnUNet \cite{isensee2021nnu} & 27.94 & 34.40 & 86.40 & 67.60 & 75.85 & 61.78 \\
  & Improved V-Net \cite{liu2018improved} & 42.43 & 49.06 & 84.27 & 58.16 & 68.82 & 51.45 \\
  & SGSNet \cite{yang2021structure} & 44.34 & 50.29 & 82.54 & 55.36 & 66.27 & 48.14 \\
  & GBP-Net \cite{yan2025glancing} & 20.35 & 28.44 & 91.13 & 76.39 & 83.11 & 71.41 \\
  & ADTL-Net \cite{liu2024brain} & 46.74 & 54.87 & 82.53 & 52.77 & 64.38 & 48.79 \\
  & Mask Transfiner \cite{ke2022mask} & 36.23 & 43.12 & 88.26 & 63.92 & 74.14 & 54.87 \\
  & SegFix \cite{yuan2020segfix} & 36.42 & 42.97 & 87.21 & 62.94 & 73.11 & 55.96 \\
  & SegRefiner \cite{wang2023segrefiner} & 30.18 & 36.25 & 88.25 & 65.19 & 74.99 & 60.32 \\
  & Ours & \textbf{16.31} & \textbf{23.66} & \textbf{92.77} & \textbf{80.38} & \textbf{86.13} & \textbf{74.25} \\

\bottomrule
\end{tabular}
}
\label{tableC}
\end{table}

\cref{tabzbf} summarizes results on the BigNeuron and ZBFWB datasets, where all refinement approaches utilize 3D U-Net segmentation outputs as initialization. Benefiting from multi-step refinement, both SegRefiner and our approach boost F1 scores on the BigNeuron dataset. Specifically, our method improves MES by 5.20\% and reduces SSD by 4.33 relative to the initial segmentation. On the more challenging ZBFWB dataset, conventional CNN-based methods struggle to handle low-contrast regions, resulting in fragmented segmentation and suppressed recall. Although GBP-Net leverages enhanced multi-scale features to surpass other methods, it remains unable to correct topological defects. Furthermore, we observe that all segmentation refinement approaches yield consistent improvements. Notably, our method elevates the F1 score of 3D U-Net by 13.19\%, achieving an F1 score 3.02\% higher than GBP-Net. These results validate the effectiveness of our multi-agent system in leveraging topological priors and global features for iterative refinement.

\begin{figure*}[!t]
\centerline{\includegraphics[width=1\columnwidth]{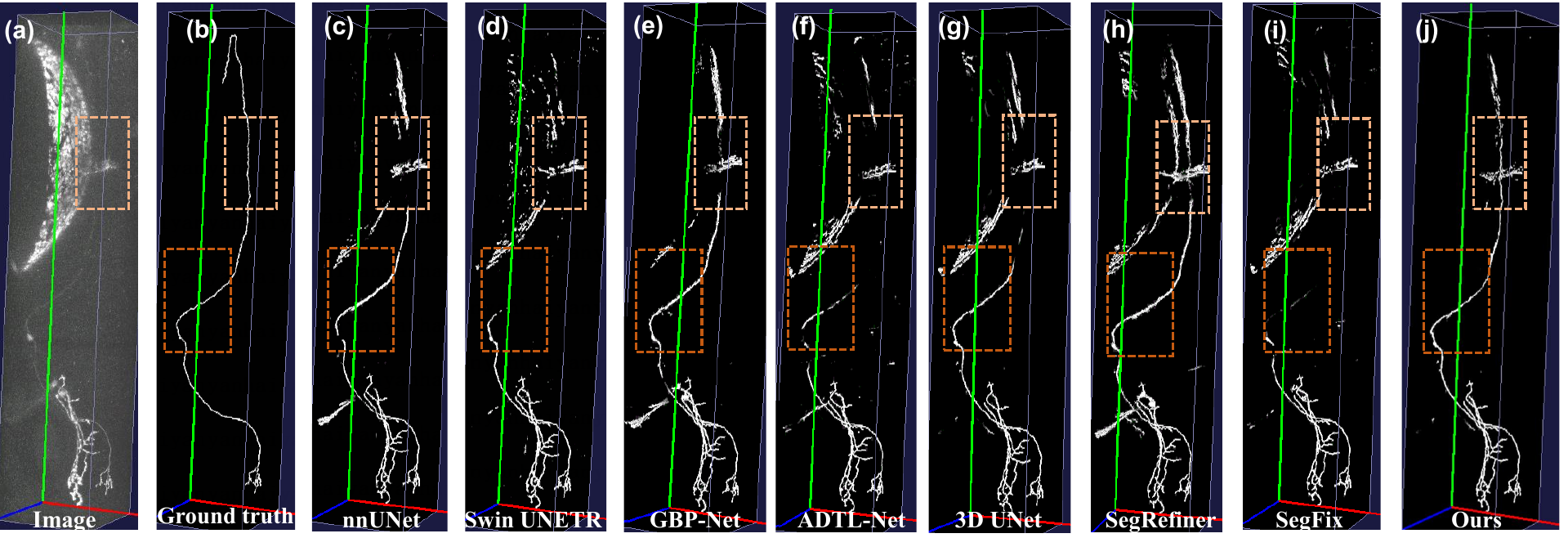}}
\caption{Qualitative comparison of original images, ground truth, and segmentation results on the ZBFWB dataset. The orange boxes highlight regions with weak signals. Our method preserves the complete neuronal structure and yields superior performance.}
\label{fig4}
\end{figure*}

\begin{table}[!t]

\caption{Ablation Study on Agents in NeuroRefiner.}
\label{MAS}
\centering
\resizebox{0.8\textwidth}{!}{

\begin{tabularx}{\textwidth}{*{3}{>{\centering\arraybackslash}X}|*{4}{>{\centering\arraybackslash}X}}
\toprule
\multirow{2}{*}{\textbf{\makecell{Global \\ Inspector}}} & 
\multirow{2}{*}{\textbf{\makecell{Refinement \\ Advisor}}} & 
\multirow{2}{*}{\textbf{\makecell{Change \\ Validator}}} & 
\multicolumn{2}{c}{\textbf{ZBFWB}} & \multicolumn{2}{c}{\textbf{CWMBS}} \\
& & &SSD & F1 (\%) & SSD & F1 (\%) \\
\midrule
&  &   & 34.52 & 76.39 & 30.23 & 79.96 \\
\checkmark &             &           &32.05&77.29&28.47 &80.34 \\
\checkmark &  \checkmark &           & 25.24 & 83.47& 18.93&87.14\\
\checkmark &            & \checkmark & 30.69 & 81.43 & 24.53 & 83.12 \\        
\midrule
\checkmark & \checkmark & \checkmark & \textbf{23.66} & \textbf{86.13} & \textbf{16.65} & \textbf{88.83} \\
\bottomrule
\end{tabularx}
}
\end{table}

\cref{fig4} depicts a neuron with long-range projections from the ZBFWB dataset, where irrelevant tissue signals significantly obscure the target structure. All single-pass segmentation methods yield results plagued by topological violations, characterized by extensive fragmentation and artifacts. Although refinement-based paradigms correct local errors in segmentation, they lack a neuron-specific error identification and correction protocol. In contrast, our method leverages agent-based reasoning and neuronal topological priors to guide the refinement with TopoRefineNet, achieving superior reconstructions.

\subsection{Ablation Study}
\subsubsection{Ablation Study on Agents in NeuroRefiner.}

To validate the efficacy of the proposed multi-agent system, we conducted ablation studies on the ZBFWB and CWMBS datasets, summarized in Table \ref{MAS}. In the absence of agents, TopoRefineNet refines all blocks indiscriminately for $T_{max}$ rounds. While this improves initial segmentation, the lack of global context and targeted guidance yields suboptimal results. Employing solely the Global Inspector for both error localization and instruction generation provides negligible gains, as a single agent struggles to manage these distinct tasks and lacks quality control mechanisms. The Refinement Advisor decouples instruction generation from inspection. This functional specialization yields F1 gains of 6.18\% and 6.80\%, respectively. Further incorporating the Change Validator to filter unreliable refinements yields additional gains of 2.66\% and 1.68\%. Ultimately, the full system achieves optimal performance, confirming that synergistic collaboration is essential for robust, high-fidelity neuron segmentation refinement.

\cref{fig:tmax_ablation} illustrates the impact of the number of iterations. The results indicate that across all three datasets, the initial iteration yields substantial gains, whereas improvements diminish significantly after the fifth iteration. Consequently, we set the maximum iteration count $T_{max}=5$. Notably, the more challenging ZBFWB and CWMBS datasets require more steps to converge and demonstrate significant improvement over the baseline segmentation.

\subsubsection{Ablation Study on Foundation models.} 
\begin{figure}[tb]
  \centering
  \begin{minipage}{0.42\linewidth}
    \centering
    \captionof{table}{Ablation Study on Foundation Model.}
    \label{tab:FM_ablation}
    \renewcommand{\arraystretch}{1.2}
    \begin{tabular}{lcccc}
    \toprule
    \multirow{2}{*}{Method} & \multicolumn{2}{c}{CWMBS} & \multicolumn{2}{c}{ZBFWB} \\
     & SSD & F1 & SSD & F1 \\
    \midrule
    Intern-VL3-8B & 17.70 & 87.94 & 28.11 & 83.81 \\
    Qwen2.5-VL-7B   & 18.33 & 87.14 & 27.31 & 84.77 \\
    Qwen3-VL-8B   & \textbf{16.65} & \textbf{88.83} & \textbf{23.66} & \textbf{86.13} \\
    \bottomrule
  \end{tabular}
  \end{minipage}
  \hfill
\begin{minipage}{0.48\linewidth}
  \centering
  \includegraphics[width=\linewidth]{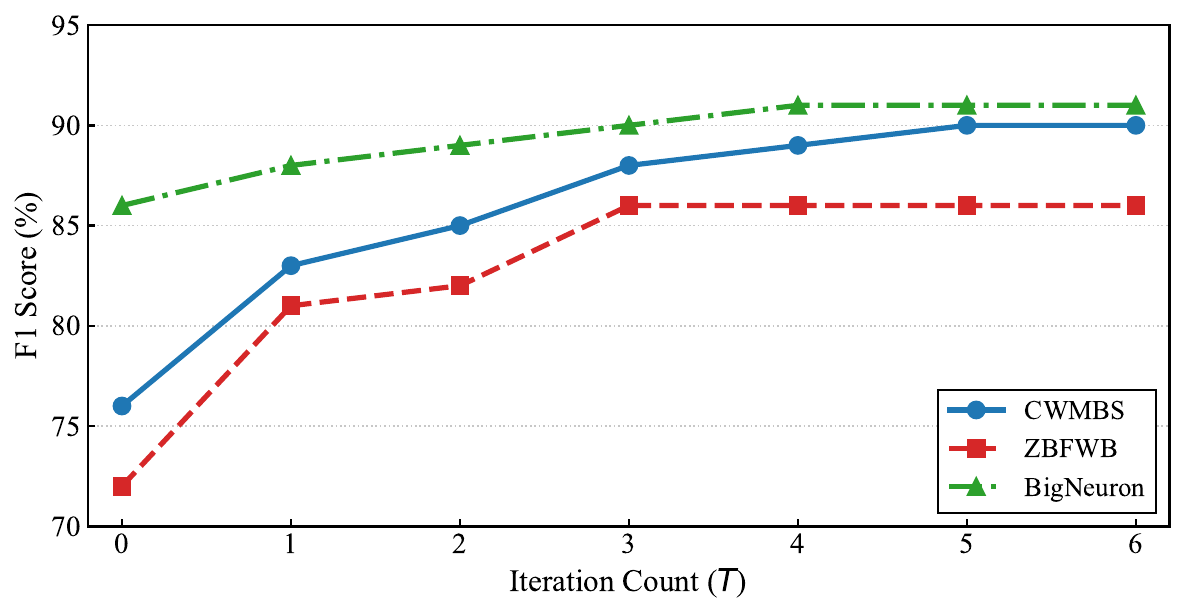}
  \caption{Ablation Study on Iteration Numbers $T_{\max}$.}
  \label{fig:tmax_ablation}
\end{minipage}
\end{figure}

We evaluated NeuroRefiner using alternative open-source VLMs. As shown in \cref{tab:FM_ablation}, while Qwen3-VL-8B achieves optimal performance due to its advanced reasoning capabilities, substituting it with weaker models like Qwen2.5-VL-7B \cite{bai2025qwen2} and Intern-VL3-8B \cite{zhu2025internvl3} still yields significant improvements over the baseline (+4.82\% F1 with Intern-VL3 on CWMBS). This confirms that the performance gain stems primarily from the multi-agent refinement framework and TopoRefineNet, rather than the specific foundation model.

\subsubsection{Ablation Study on Global Inspector.} We conduct ablation studies on the input configuration, as summarized in \cref{GI}. When relying solely on \(M^{xy}\), the inspector tends to overlook fragmentation artifacts, leading to marginal refinement gains. In contrast, providing the full sequence of patch-level projections alongside the global view enables it to jointly leverage holistic context and local cues. This significantly enhances the ability to pinpoint topological defects and improves the F1 score by 3.43\% and 3.20\%. Although the MIP of the segmentation results is sufficient to identify most topological anomalies in sparsely distributed neurons, a small number of complex volumes may still suffer from interference due to neuron overlap.
Therefore, incorporating the 0-th Betti number \(\beta_0\) as an auxiliary topological indicator effectively directs attention toward regions with high connected component counts. This configuration achieved the best performance across multiple datasets.

\begin{table}[!t]
\centering

\begin{minipage}[t]{0.55\linewidth} 
    \centering
    \captionof{table}{Ablation Study on Global Inspector.}
    \label{GI}
    \renewcommand{\arraystretch}{0.9} 
    \resizebox{1.0\linewidth}{!}{
    \begin{tabular}{ccccccc}
    \toprule
    \multicolumn{3}{c|}{Inputs} & \multicolumn{2}{c}{ZBFWB} & \multicolumn{2}{c}{CWMBS} \\
    $M^{xy}$ & $\{M_i^{\mathrm{xy}}\}$ & ${\beta_0(b_i)}$ & SSD & F1 (\%) & SSD & F1 (\%) \\
    \midrule
    \checkmark &  &            & 32.86 & 77.36 & 27.11 & 81.89 \\
    \checkmark &  \checkmark&  & 26.27 & 80.79 & 21.47 & 85.09 \\
    \checkmark &  & \checkmark & 29.16 & 82.41 & 25.93 & 82.64 \\
    \midrule
    \checkmark & \checkmark & \checkmark & \textbf{23.66} & \textbf{86.13} & \textbf{16.65} & \textbf{88.83}\\
    \bottomrule
    \end{tabular}
    }
\end{minipage}%
\hfill
\begin{minipage}[t]{0.4\linewidth} 
    \centering
    \captionof{table}{Ablating TopoRefineNet.}
    \label{tab:fusion_ablation}
    \renewcommand{\arraystretch}{1.2}
    \resizebox{1\textwidth}{!}{
    \begin{tabular}{lcccc}
    \toprule
    \multirow{2}{*}{Method} & \multicolumn{2}{c}{CWMBS} & \multicolumn{2}{c}{ZBFWB} \\
     & SSD & F1 & SSD & F1 \\
    \midrule
    AdaLN & 19.37 & 86.46 & 26.78 & 84.38 \\
    CDP   & 17.89 & 88.07 & \textbf{22.14} & 86.07 \\
    CAB   & \textbf{16.65} & \textbf{88.83} & 23.66 & \textbf{86.13} \\
    \bottomrule
    \end{tabular}
    }
\end{minipage}

\end{table}

\subsubsection{Ablation Study on TopoRefineNet.}
We carry out ablation studies on the architecture and training strategies of TopoRefineNet. First, we compared three multimodal fusion strategies: channel-wise dot product (CDP) \cite{rokuss2025voxtell}, cross-attention in the bottleneck (CAB), and AdaLN \cite{peebles2023scalable}. As shown in \cref{tab:fusion_ablation}, since CDP and CAB yield comparable accuracy, we adopt the more efficient CAB. Second, we replace the text encoder with the larger Qwen3-4B. It yields no significant improvement, indicating that the 0.6B variant suffices for precise encoding of the structured instructions. To validate the two-stage curriculum learning strategy, we trained on real defects from scratch as a comparison. This single-stage variant exhibited a 23.57\% increase in the rejection rate of Change Validator, alongside F1 score degradations of 1.17\% and 1.89\% in ZBFWB and CWMBS, respectively. The decline confirms the necessity of our training strategy.

\section{Conclusion}

In this paper, we present NeuroRefiner, the first LLM-based multi-agent system for neuron segmentation refinement. By coordinating morphology-aware global inspection, language-guided editing, and validation, our approach achieves progressive refinement without agent training. Consequently, it overcomes the limited topological fidelity and poor interpretability inherent in end-to-end segmentation models. The proposed TopoRefineNet bridges the gap between instructions and segmentation models, offering novel insights into agent–tool interaction design. Notably, our framework is agnostic to the foundation model, ensuring it evolves in tandem with advancements in VLMs.
\vspace{-0.3em}
\subsubsection{Limitation}
While NeuroRefiner's 2D MIP detects topological anomalies in sparse neurons, it loses depth information in dense regions, hindering precise 3D error localization. Future work can introduce depth-aware encoding, which maps Z-axis depth to pseudo-color channels or generates depth-weighted projections, allowing the Global Inspector to resolve occlusions and accurately infer the Z-axis depth of topological defects.
\vspace{-0.3em}
\subsubsection{Acknowledgement}
This work was supported by the Brain Science and Brain-like Intelligence Technology - National Science and Technology Major Project (2021ZD0204500, 2021ZD0204503 to L.L.), National Key Research and Development Program of China (2025YFA1614600 to J.L.).


%
%
\bibliographystyle{splncs04}
\bibliography{main}

@article{gou2024gapr,
  title={Gapr for large-scale collaborative single-neuron reconstruction},
  author={Gou, Lingfeng and Wang, Yanzhi and Gao, Le and Zhong, Yiting and Xie, Lucheng and Wang, Haifang and Zha, Xi and Shao, Yinqi and Xu, Huatai and Xu, Xiaohong and others},
  journal={Nature Methods},
  volume={21},
  number={10},
  pages={1926--1935},
  year={2024},
  publisher={Nature Publishing Group US New York}
}

@article{zhang2023collaborative,
  title={Collaborative Augmented Reconstruction for Scaled Production of 3D Neuron Morphology in Mouse and Human Brains},
  author={Zhang, Lingli and Huang, Lei and Yuan, Zexin and Hang, Yuning and Zeng, Ying and Li, Kaixiang and Wang, Lijun and Zeng, Haoyu and Chen, Xin and Zhang, Hairuo and others},
  journal={bioRxiv},
  pages={2023--10},
  year={2023},
  publisher={Cold Spring Harbor Laboratory}
}

@article{du2025central,
  title={Central nervous system atlas of larval zebrafish constructed using the morphology of single excitatory and inhibitory neurons},
  author={Du, Xufei and Yue, Zhifeng and Wei, Jiafei and Li, Wanlan and Chen, Mingquan and Chen, Tianlun and Hu, Hanyang and Ren, Hongan and Jia, Zhiming and Ning, Xinyu and others},
  journal={bioRxiv},
  pages={2025--06},
  year={2025},
  publisher={Cold Spring Harbor Laboratory}
}

@article{xiao2013app2,
  title={APP2: Automatic tracing of {3D} neuron morphology based on hierarchical pruning of a gray-weighted image distance-tree},
  author={Xiao, Hang and Peng, Hanchuan},
  journal={Bioinformatics},
  volume={29},
  number={11},
  pages={1448--1454},
  year={2013},
  publisher={Oxford University Press}
}

@inproceedings{milletari2016v,
  title={V-net: Fully convolutional neural networks for volumetric medical image segmentation},
  author={Milletari, Fausto and Navab, Nassir and Ahmadi, Seyed-Ahmad},
  booktitle={Proc. 2016 4th Int. Conf. {3D} Vis. (3DV)},
  pages={565--571},
  year={2016},
  organization={IEEE}
}

@article{yang2021structure,
  title={Structure-guided segmentation for {3D} neuron reconstruction},
  author={Yang, Bo and Liu, Min and Wang, Yaonan and Zhang, Kang and Meijering, Erik},
  journal={IEEE Trans. Med. Imag.},
  volume={41},
  number={4},
  pages={903--914},
  year={2021},
  publisher={IEEE}
}

@article{wang2023nrtr,
  title={NRTR: Neuron reconstruction with transformer from {3D} optical microscopy images},
  author={Wang, Yijun and Lang, Rui and Li, Rui and Zhang, Junsong},
  journal={IEEE Trans. Med. Imag.},
  year={2023},
  publisher={IEEE}
}

@article{chen2023deep,
  title={Deep learning in mesoscale brain microscopy image analysis: A review},
  author={Chen, Runze and Liu, Min and Chen, Weixun and Wang, Yaonan and Meijering, Erik},
  journal={Computers in Biology and Medicine},
  pages={107617},
  year={2023},
  publisher={Elsevier}
}

@article{liu2022neuron,
  title={Neuron tracing from light microscopy images: Automation, deep learning and bench testing},
  author={Liu, Yufeng and Wang, Gaoyu and Ascoli, Giorgio A and Zhou, Jiangning and Liu, Lijuan},
  journal={Bioinformatics},
  volume={38},
  number={24},
  pages={5329--5339},
  year={2022},
  publisher={Oxford University Press}
}

@inproceedings{liu2025netracer,
  title={NETracer: A Topology-Aware Iterative Tracing Approach for Tubular Structure Extraction},
  author={Liu, Chao and Jiang, Yangbo and Zheng, Nenggan},
  booktitle={Proceedings of the IEEE/CVF International Conference on Computer Vision},
  pages={20593--20602},
  year={2025}
}

@article{ma2025medsam2,
  title={Medsam2: Segment anything in 3d medical images and videos},
  author={Ma, Jun and Yang, Zongxin and Kim, Sumin and Chen, Bihui and Baharoon, Mohammed and Fallahpour, Adibvafa and Asakereh, Reza and Lyu, Hongwei and Wang, Bo},
  journal={arXiv preprint arXiv:2504.03600},
  year={2025}
}

@article{chen2021weakly,
  title={Weakly supervised neuron reconstruction from optical microscopy images with morphological priors},
  author={Chen, Xuejin and Zhang, Chi and Zhao, Jie and Xiong, Zhiwei and Zha, Zheng-Jun and Wu, Feng},
  journal={IEEE Trans. Med. Imag.},
  volume={40},
  number={11},
  pages={3205--3216},
  year={2021},
  publisher={IEEE}
}

@article{wang20213d,
  title={A {3D} tubular flux model for centerline extraction in neuron volumetric images},
  author={Wang, Xuan and Liu, Min and Wang, Yaonan and Fan, Jiawang and Meijering, Erik},
  journal={IEEE Trans. Med. Imag.},
  volume={41},
  number={5},
  pages={1069--1079},
  year={2021},
  publisher={IEEE}
}

@inproceedings{wang2021single,
  title={Single neuron segmentation using graph-based global reasoning with auxiliary skeleton loss from {3D} optical microscope images},
  author={Wang, Heng and Song, Yang and Zhang, Chaoyi and Yu, Jianhui and Liu, Siqi and Pengy, Hanchuan and Cai, Weidong},
  booktitle={IEEE 18th Int. Symp. Biomed. Imaging (ISBI)},
  pages={934--938},
  year={2021},
  organization={IEEE}
}

@inproceedings{zhao2023pointneuron,
  title={PointNeuron: 3D neuron reconstruction via geometry and topology learning of point clouds},
  author={Zhao, Runkai and Wang, Heng and Zhang, Chaoyi and Cai, Weidong},
  booktitle={Proceedings of the IEEE/CVF winter conference on applications of computer vision},
  pages={5787--5797},
  year={2023}
}

@article{2019Learning,
  title={Learning Shape Representation on Sparse Point Clouds for Volumetric Image Segmentation},
  author={ Balsiger, Fabian  and  Soom, Yannick  and  Scheidegger, Olivier  and  Reyes, Mauricio },
  year={2019},
}

@article{yang2020neuron,
  title={Neuron image segmentation via learning deep features and enhancing weak neuronal structures},
  author={Yang, Bo and Chen, Weixun and Luo, Huiqiong and Tan, Yinghui and Liu, Min and Wang, Yaonan},
  journal={IEEE J. Biomed. Health Informat.},
  volume={25},
  number={5},
  pages={1634--1645},
  year={2020},
  publisher={IEEE}
}

@article{peng2015bigneuron,
  title={BigNeuron: Large-scale {3D} neuron reconstruction from optical microscopy images},
  author={Peng, Hanchuan and Hawrylycz, Michael and Roskams, Jane and Hill, Sean and Spruston, Nelson and Meijering, Erik and Ascoli, Giorgio A},
  journal={Neuron},
  volume={87},
  number={2},
  pages={252--256},
  year={2015},
  publisher={Elsevier}
}

@inproceedings{yan2024neurolink,
  title={NeuroLink: Bridging Weak Signals in Neuronal Imaging with Morphology Learning},
  author={Yan, Haiyang and Zhai, Hao and Guo, Jinyue and Li, Linlin and Han, Hua},
  booktitle={Int. Conf. Med. Image Comput. Comput.-Assist. Interv. (MICCAI)},
  pages={467--477},
  year={2024},
  organization={Springer}
}

@article{yu2025gencellagent,
  title={GenCellAgent: Generalizable, Training-Free Cellular Image Segmentation via Large Language Model Agents},
  author={Yu, Xi and Yang, Yang and Liu, Qun and Du, Yonghua and McSweeney, Sean and Lin, Yuewei},
  journal={arXiv preprint arXiv:2510.13896},
  year={2025}
}

@article{jiang2026ibisagent,
  title={IBISAgent: Reinforcing Pixel-Level Visual Reasoning in MLLMs for Universal Biomedical Object Referring and Segmentation},
  author={Jiang, Yankai and Li, Qiaoru and Xu, Binlu and Sun, Haoran and Ding, Chao and Dong, Junting and Cai, Yuxiang and Zhang, Xuhong and Yin, Jianwei},
  journal={arXiv preprint arXiv:2601.03054},
  year={2026}
}

@article{jiang2025incentivizing,
  title={Incentivizing Tool-augmented Thinking with Images for Medical Image Analysis},
  author={Jiang, Yankai and Zhang, Yujie and Zhang, Peng and Li, Yichen and Chen, Jintai and Shi, Xiaoming and Zhen, Shihui},
  journal={arXiv preprint arXiv:2512.14157},
  year={2025}
}

@article{bai2025qwen3,
  title={Qwen3-vl technical report},
  author={Bai, Shuai and Cai, Yuxuan and Chen, Ruizhe and Chen, Keqin and Chen, Xionghui and Cheng, Zesen and Deng, Lianghao and Ding, Wei and Gao, Chang and Ge, Chunjiang and others},
  journal={arXiv preprint arXiv:2511.21631},
  year={2025}
}

@article{zhang2025qwen3,
  title={Qwen3 embedding: Advancing text embedding and reranking through foundation models},
  author={Zhang, Yanzhao and Li, Mingxin and Long, Dingkun and Zhang, Xin and Lin, Huan and Yang, Baosong and Xie, Pengjun and Yang, An and Liu, Dayiheng and Lin, Junyang and others},
  journal={arXiv preprint arXiv:2506.05176},
  year={2025}
}

@article{2024BiomedParse,
  title={BiomedParse: a biomedical foundation model for image parsing of everything everywhere all at once},
  author={ Zhao, Theodore  and  Gu, Yu  and  Yang, Jianwei  and  Usuyama, Naoto  and  Lee, Ho Hin  and  Naumann, Tristan  and  Gao, Jianfeng  and  Crabtree, Angela  and  Abel, Jacob  and  Moung-Wen, Christine },
  year={2024},
}

@article{2024SAM,
  title={SAM 2: Segment Anything in Images and Videos},
  author={ Ravi, Nikhila  and  Gabeur, Valentin  and  Hu, Yuan Ting  and  Hu, Ronghang  and  Ryali, Chaitanya  and  Ma, Tengyu  and  Khedr, Haitham  and  Rdle, Roman  and  Rolland, Chloe  and  Gustafson, Laura },
  year={2024},
}

@inproceedings{ke2022mask,
  title={Mask transfiner for high-quality instance segmentation},
  author={Ke, Lei and Danelljan, Martin and Li, Xia and Tai, Yu-Wing and Tang, Chi-Keung and Yu, Fisher},
  booktitle={Proceedings of the IEEE/CVF Conference on Computer Vision and Pattern Recognition},
  pages={4412--4421},
  year={2022}
}

@inproceedings{yuan2020segfix,
  title={Segfix: Model-agnostic boundary refinement for segmentation},
  author={Yuan, Yuhui and Xie, Jingyi and Chen, Xilin and Wang, Jingdong},
  booktitle={European conference on computer vision},
  pages={489--506},
  year={2020},
  organization={Springer}
}

@article{wang2023segrefiner,
  title={Segrefiner: Towards model-agnostic segmentation refinement with discrete diffusion process},
  author={Wang, Mengyu and Ding, Henghui and Liew, Jun Hao and Liu, Jiajun and Zhao, Yao and Wei, Yunchao},
  journal={arXiv preprint arXiv:2312.12425},
  year={2023}
}

@article{peng2010v3d,
  title={V3D enables real-time {3D} visualization and quantitative analysis of large-scale biological image data sets},
  author={Peng, Hanchuan and Ruan, Zongcai and Long, Fuhui and Simpson, Julie H and Myers, Eugene W},
  journal={Nature Biotechnol.},
  volume={28},
  number={4},
  pages={348--353},
  year={2010},
  publisher={Nature Publishing Group}
}

@article{xie2011anisotropic,
  title={Anisotropic path searching for automatic neuron reconstruction},
  author={Xie, Jun and Zhao, Ting and Lee, Tzumin and Myers, Eugene and Peng, Hanchuan},
  journal={Med. Image Anal.},
  volume={15},
  number={5},
  pages={680--689},
  year={2011},
  publisher={Elsevier}
}

@inproceedings{zhang2023adding,
  title={Adding conditional control to text-to-image diffusion models},
  author={Zhang, Lvmin and Rao, Anyi and Agrawala, Maneesh},
  booktitle={Proceedings of the IEEE/CVF international conference on computer vision},
  pages={3836--3847},
  year={2023}
}

@inproceedings{rombach2022high,
  title={High-resolution image synthesis with latent diffusion models},
  author={Rombach, Robin and Blattmann, Andreas and Lorenz, Dominik and Esser, Patrick and Ommer, Bj{\"o}rn},
  booktitle={Proceedings of the IEEE/CVF conference on computer vision and pattern recognition},
  pages={10684--10695},
  year={2022}
}

@article{couairon2022diffedit,
  title={Diffedit: Diffusion-based semantic image editing with mask guidance},
  author={Couairon, Guillaume and Verbeek, Jakob and Schwenk, Holger and Cord, Matthieu},
  journal={arXiv preprint arXiv:2210.11427},
  year={2022}
}

@article{yan2025glancing,
  title={Glancing Beyond Patch: Spatial Contextual Cues for 3D Neuron Segmentation},
  author={Yan, Haiyang and Zhang, Yanchao and Li, Zhenchen and Guo, Jinyue and Zhai, Hao and Liu, Jiazheng and Zhong, Yongwei and Yuan, Jingbin and Shen, Lijun and Li, Linlin and others},
  journal={IEEE Transactions on Medical Imaging},
  year={2025},
  publisher={IEEE}
}

@article{liu2024brain,
  title={Brain Image Segmentation for Ultrascale Neuron Reconstruction via an Adaptive Dual-Task Learning Network},
  author={Liu, Min and Wu, Shuhan and Chen, Runze and Lin, Zhuangdian and Wang, Yaonan and Meijering, Erik},
  journal={IEEE Trans. Med. Imag.},
  year={2024},
  publisher={IEEE}
}

@inproceedings{cciccek20163d,
  title={3D U-Net: learning dense volumetric segmentation from sparse annotation},
  author={{\c{C}}i{\c{c}}ek, {\"O}zg{\"u}n and Abdulkadir, Ahmed and Lienkamp, Soeren S and Brox, Thomas and Ronneberger, Olaf},
  booktitle={International conference on medical image computing and computer-assisted intervention},
  pages={424--432},
  year={2016},
  organization={Springer}
}

@inproceedings{hatamizadeh2021swin,
  title={Swin unetr: Swin transformers for semantic segmentation of brain tumors in mri images},
  author={Hatamizadeh, Ali and Nath, Vishwesh and Tang, Yucheng and Yang, Dong and Roth, Holger R and Xu, Daguang},
  booktitle={International MICCAI brainlesion workshop},
  pages={272--284},
  year={2021},
  organization={Springer}
}

@inproceedings{hatamizadeh2022unetr,
  title={Unetr: Transformers for 3d medical image segmentation},
  author={Hatamizadeh, Ali and Tang, Yucheng and Nath, Vishwesh and Yang, Dong and Myronenko, Andriy and Landman, Bennett and Roth, Holger R and Xu, Daguang},
  booktitle={Proceedings of the IEEE/CVF winter conference on applications of computer vision},
  pages={574--584},
  year={2022}
}

@article{zhou2023nnformer,
  title={nnformer: Volumetric medical image segmentation via a 3d transformer},
  author={Zhou, Hong-Yu and Guo, Jiansen and Zhang, Yinghao and Han, Xiaoguang and Yu, Lequan and Wang, Liansheng and Yu, Yizhou},
  journal={IEEE transactions on image processing},
  volume={32},
  pages={4036--4045},
  year={2023},
  publisher={IEEE}
}

@inproceedings{liu2018improved,
  title={Improved V-Net based image segmentation for {3D} neuron reconstruction},
  author={Liu, Min and Luo, Huiqiong and Tan, Yinghui and Wang, Xueping and Chen, Weixun},
  booktitle={Proc. 2018 IEEE Int. Conf. Bioinf. Biomed. (BIBM)},
  pages={443--448},
  year={2018},
  organization={IEEE}
}

@article{isensee2021nnu,
  title={{nnU-Net}: A self-configuring method for deep learning-based biomedical image segmentation},
  author={Isensee, Fabian and Jaeger, Paul F and Kohl, Simon AA and Petersen, Jens and Maier-Hein, Klaus H},
  journal={Nature Methods},
  volume={18},
  number={2},
  pages={203--211},
  year={2021},
  publisher={Nature Publishing Group}
}

@inproceedings{peebles2023scalable,
  title={Scalable diffusion models with transformers},
  author={Peebles, William and Xie, Saining},
  booktitle={Proceedings of the IEEE/CVF international conference on computer vision},
  pages={4195--4205},
  year={2023}
}

@article{rokuss2025voxtell,
  title={Voxtell: Free-text promptable universal 3d medical image segmentation},
  author={Rokuss, Maximilian and Langenberg, Moritz and Kirchhoff, Yannick and Isensee, Fabian and Hamm, Benjamin and Ulrich, Constantin and Regnery, Sebastian and Bauer, Lukas and Katsigiannopulos, Efthimios and Norajitra, Tobias and others},
  journal={arXiv preprint arXiv:2511.11450},
  year={2025}
}

@article{bai2025qwen2,
  title={Qwen2. 5-vl technical report},
  author={Bai, Shuai and Chen, Keqin and Liu, Xuejing and Wang, Jialin and Ge, Wenbin and Song, Sibo and Dang, Kai and Wang, Peng and Wang, Shijie and Tang, Jun and others},
  journal={arXiv preprint arXiv:2502.13923},
  year={2025}
}

@article{zhu2025internvl3,
  title={Internvl3: Exploring advanced training and test-time recipes for open-source multimodal models},
  author={Zhu, Jinguo and Wang, Weiyun and Chen, Zhe and Liu, Zhaoyang and Ye, Shenglong and Gu, Lixin and Tian, Hao and Duan, Yuchen and Su, Weijie and Shao, Jie and others},
  journal={arXiv preprint arXiv:2504.10479},
  year={2025}
}

@article{chen2023self,
  title={Self-supervised neuron segmentation with multi-agent reinforcement learning},
  author={Chen, Yinda and Huang, Wei and Zhou, Shenglong and Chen, Qi and Xiong, Zhiwei},
  journal={arXiv preprint arXiv:2310.04148},
  year={2023}
}

@article{ma2020boundary,
  title={Boundary-aware supervoxel-level iteratively refined interactive 3D image segmentation with multi-agent reinforcement learning},
  author={Ma, Chaofan and Xu, Qisen and Wang, Xiangfeng and Jin, Bo and Zhang, Xiaoyun and Wang, Yanfeng and Zhang, Ya},
  journal={IEEE Transactions on Medical Imaging},
  volume={40},
  number={10},
  pages={2563--2574},
  year={2020},
  publisher={IEEE}
}

@inproceedings{liao2020iteratively,
  title={Iteratively-refined interactive 3D medical image segmentation with multi-agent reinforcement learning},
  author={Liao, Xuan and Li, Wenhao and Xu, Qisen and Wang, Xiangfeng and Jin, Bo and Zhang, Xiaoyun and Wang, Yanfeng and Zhang, Ya},
  booktitle={Proceedings of the IEEE/CVF conference on computer vision and pattern recognition},
  pages={9394--9402},
  year={2020}
}

@article{jiang2026neuromamba,
  title={NeuroMamba: Multi-Perspective Feature Interaction with Visual Mamba for Neuron Segmentation},
  author={Jiang, Liuyun and Lu, Yizhuo and Zhang, Yanchao and Liu, Jiazheng and Han, Hua},
  journal={arXiv preprint arXiv:2601.15929},
  year={2026}
}

@inproceedings{zhang2024segneuron,
  title={Segneuron: 3d neuron instance segmentation in any em volume with a generalist model},
  author={Zhang, Yanchao and Guo, Jinyue and Zhai, Hao and Liu, Jing and Han, Hua},
  booktitle={International Conference on Medical Image Computing and Computer-Assisted Intervention},
  pages={589--600},
  year={2024},
  organization={Springer}
}

@article{sheridan2023local,
  title={Local shape descriptors for neuron segmentation},
  author={Sheridan, Arlo and Nguyen, Tri M and Deb, Diptodip and Lee, Wei-Chung Allen and Saalfeld, Stephan and Turaga, Srinivas C and Manor, Uri and Funke, Jan},
  journal={Nature methods},
  volume={20},
  number={2},
  pages={295--303},
  year={2023},
  publisher={Nature Publishing Group US New York}
}

@article{zhang2023towards,
  title={Towards segment anything model (sam) for medical image segmentation: a survey},
  author={Zhang, Yichi and Jiao, Rushi},
  journal={arXiv preprint arXiv:2305.03678},
  year={2023}
}

@article{li2025fgnet,
  title={FGNet: Leveraging Feature-Guided Attention to Refine SAM2 for 3D EM Neuron Segmentation},
  author={Li, Zhenghua and Chen, Hang and Sun, Zihao and Li, Kai and Hu, Xiaolin},
  journal={arXiv preprint arXiv:2511.13063},
  year={2025}
}

@article{ravi2024sam,
  title={Sam 2: Segment anything in images and videos},
  author={Ravi, Nikhila and Gabeur, Valentin and Hu, Yuan-Ting and Hu, Ronghang and Ryali, Chaitanya and Ma, Tengyu and Khedr, Haitham and R{\"a}dle, Roman and Rolland, Chloe and Gustafson, Laura and others},
  journal={arXiv preprint arXiv:2408.00714},
  year={2024}
}

\begin{figure*}[h]
\renewcommand{\thefigure}{S\arabic{figure}}
  \center
\begin{tcolorbox}[colback=gray!5!white,colframe=gray!75!black,title=Prompt for Global Inspector,top=0pt,bottom=0pt]
\begin{lstlisting}[
    basicstyle=\ttfamily\scriptsize,
    breaklines=true,  % 允许自动换行
    breakatwhitespace=true,
    frame=none,
    abovecaptionskip=5pt, 
]
You are a Neuron Segmentation Topology Quality Inspection Expert. Evaluate segmentation quality based on the morphological features of the Segmentation Mask and Connected Components (CC) metrics.

# Task
You will receive a set of neuron segmentation data:
1. **Global Segmentation Map**: The overall segmentation result marked with Block IDs.
2. **Sub-region Masks**: Binary segmentation masks for each Block.
3. **Connected Components Metrics**: The pre-calculated number of Connected Components within each Block.

Your goal is to identify erroneous blocks that cause neuron discontinuity or contain noise. The final ideal state is: **Only a few connected components across the entire map, with no isolated noise.**

# Judgment Logic 
Please make judgments based on the following morphological and topological rules:

## 1. False Negative (FN) - Topological Breakage
- **Characteristic**: Neurons should be continuous but are disconnected within the block or at block boundaries.
- **Mask Appearance**: 
  - Multiple large connected components exist within the block, oriented towards each other but unconnected.
  - The neuron skeleton abruptly terminates within the block (not at the boundary).
  - Neuron endings at the block boundary do not find continuation in adjacent blocks.

## 2. False Positive (FP) - Isolated Noise
- **Characteristic**: Background areas incorrectly labeled as neurons.
- **Mask Appearance**: 
  - Presence of tiny, isolated connected components (spot-like).
  - Component shapes are irregular and do not conform to the neuron "slender tubular" morphological prior.
  - No physical connection with the main neuron skeleton.

## 3. Normal: Good connectivity within the block, morphology conforms to neuron characteristics.

# Workflow
1. **Metric Filtering**: Prioritize Blocks with abnormally high CC counts, as this usually indicates breakage or noise.
2. **Morphological Analysis**: Observe the shapes of components within high-CC blocks. Is it FN or FP?
3. **Boundary Check**: Observe whether the segmentation at block edges is smooth and continuous, or if there are abrupt truncations.
4. **Global Consistency**: Consider whether correcting this block would help improve the connectivity of the entire map.

# Output Format
Only output problematic Block IDs. Values are limited to "FN" or "FP".
Format Example:
{"1": "FN", "5": "FP", "12": "FN"}

# Constraints
- **Do Not Fabricate**: Judge solely based on the provided mask morphology; do not assume raw signals that do not exist.
- **Strict Format**: Must comply with JSON standards.
- **Convergence Condition**: If the neuron morphology is judged to be intact, output an empty JSON object.
\end{lstlisting}
\end{tcolorbox}
\label{RA}
\end{figure*}

\begin{figure*}[h]
\renewcommand{\thefigure}{S\arabic{figure}}
  \center
\begin{tcolorbox}[colback=gray!5!white,colframe=gray!75!black,title=Prompt for Refinement Advisor, top=0pt,bottom=0pt]
\begin{lstlisting}[
    basicstyle=\ttfamily\scriptsize,
    breaklines=true,  % 允许自动换行
    breakatwhitespace=true,
    frame=none,
    abovecaptionskip=5pt, 
]
You are a senior biomedical image analysis expert specializing in the reconstruction and segmentation quality assessment of 3D neuronal microscopy images. You possess strong spatial reasoning capabilities and can infer 3D structures from 2D Maximum Intensity Projection (MIP) images.

You will receive 4 images, categorized into two view groups (XY plane and YZ plane). Each group contains one "Original Fluorescence Image" and one "Segmentation Mask". The defects present in the current segmentation results are: [Placeholder]. Specifically:
  - **False Negative (FN)**: Signal exists in the original image, but is missing in the segmentation result (broken or lost).
  - **False Positive (FP)**: No signal in the original image, but marked in the segmentation result (noise or over-segmentation).
Your core task is to locate errors across dual views by comparing the original image with the segmentation results, and generate executable structured correction instructions.

# Input Definition
Image 1: Original image in XY view. Image 2: Segmentation result in XY view. Image 3: Original image in YZ view. 4. Image 4: Segmentation result in YZ view.

Please strictly follow the steps below for reasoning and output:

## Step 1: Compare Image 1 and Image 2.
- **Clues**: Identify corresponding fluorescence signal features (e.g., brightness, continuity, texture) in the original image (Image 1). For False Negatives, look for potential connection clues. For False Positives, identify the location of noise.
- **Localization**: Determine the relative position of the error on the XY plane (Top/Bottom/Left/Right/Center).

## Step 2: YZ View Analysis and Localization
- Compare Image 3 and Image 4 to determine the relative position of the error on the YZ plane (Top/Bottom/Front/Rear/Center). *Note: The left side represents the Front direction*.

## Step 3: Generate Correction Instructions
Based on the error classification, generate standardized natural language instructions:
- **For False Negatives (FN)**:
  - **Long-range Discontinuity**: If the neuron trunk break spans a large distance, specify connection endpoints. Format: `Connect neuron from [Start Position] to [End Position]` (Example: Connect neuron from top-left-rear to bottom-left-front).
  - **Short-range Discontinuity**: If it is only a local gap, specify the break point. Format: `Repair break at [Specific Position]` (Example: Repair break at top-right-front).
- **For False Positives (FP)**: Specify the region to be removed. Format: `Remove artifact/noise at [Specific Position]` (Example: Remove isolated noise at right-center-rear) or `Remove noise from [Start Position] to [End Position]`.

# Constraints
- Must be based on image evidence; fabricating non-existent structures is strictly prohibited.
- Position descriptions must include both planar information (e.g., "Top-Left") and depth information (e.g., "Rear") to ensure accurate 3D localization.
- Output must be objective and concise; avoid ambiguous vocabulary (e.g., "approximately", "possibly").
\end{lstlisting}
\end{tcolorbox}
\label{SA}
\end{figure*}

\begin{figure*}[h]
\renewcommand{\thefigure}{S\arabic{figure}} 
  \center
\begin{tcolorbox}[colback=gray!5!white,colframe=gray!75!black,title=Prompt for Change Validator,top=0pt,bottom=0pt]
\begin{lstlisting}[
    basicstyle=\ttfamily\scriptsize,
    breaklines=true,  % 允许自动换行
    breakatwhitespace=true,
    frame=none,
    abovecaptionskip=5pt, 
]
You are a senior Neuron Segmentation Quality Assessment Specialist. Your task is to evaluate the effectiveness of automated correction workflows on neuron segmentation results.

# Input Data
1. **Image A (Pre-correction)**: Initial neuron segmentation result.
2. **Image B (Post-correction)**: Segmentation result after algorithmic processing.
3. **Correction Command**: {{correction_command_placeholder}}

# Evaluation Criteria
Compare Image A and Image B strictly based on the following two dimensions. **The correction is accepted ONLY if BOTH dimensions are satisfied.**

## Dimension 1: Topological Quality
Image B must demonstrate equal or superior structural integrity compared to Image A. Focus on:
- **Continuity**: Are neurites more continuous with fewer breaks/discontinuities?
- **Noise Control**: Are background artifacts or isolated noise points effectively suppressed?
- **Connectivity**: Are critical branching points preserved with plausible connections, avoiding erroneous disconnections or abnormal mergers?
*Note: Perfection is not required, but a clear trend toward "denoising" or "connection repair" must be observable.*

## Dimension 2: Instruction Compliance
Changes in Image B must explicitly align with the semantic intent of the [Correction Command].
- If the command is "connect broken segments", is the specified break actually connected in Image B?
- If the command is "remove noise", are the targeted artifacts eliminated in Image B?
- If the command is ignored, partially executed, or misapplied (e.g., valid neurites erroneously deleted), mark as non-compliant.

# Workflow
1. **Visual Comparison**: Carefully examine differences between Image A and B.
2. **Command Verification**: Semantically match observed changes against the [Correction Command].
3. **Topology Assessment**: Judge whether structural quality improved (cleaner, more continuous).
4. **Final Decision**: Apply logic: `Accept = (Topology Improved OR Unchanged) AND (Command Compliant)`.

**Evaluation Conclusion**:
- Result: [ACCEPT / REJECT]
- Reason: [If REJECTED, specify whether due to topology degradation or command non-compliance; if ACCEPTED, briefly state the improvement]
\end{lstlisting}
\end{tcolorbox}
\label{VPA}
\end{figure*}

\end{document}